\documentclass[journal]{IEEEtran}
\usepackage{graphicx}
\usepackage{cite}
\usepackage{picinpar}
\usepackage{amsmath}
\usepackage{url}
\usepackage{flushend}
\usepackage[latin1]{inputenc}
\usepackage{colortbl}
\usepackage{soul}
\usepackage{multirow}
\usepackage{pifont}
\usepackage{alltt}
\usepackage[hidelinks]{hyperref}
\usepackage{enumerate}
\usepackage{siunitx}
\usepackage{breakurl}
\usepackage{epstopdf}
\usepackage{pbox}

\usepackage{amsmath,amsfonts}
\usepackage{array}
\usepackage[caption=false,font=normalsize,labelfont=sf,textfont=sf]{subfig}
\usepackage{textcomp}
\usepackage{stfloats}
\usepackage{verbatim}
\usepackage{lineno}
\usepackage{amssymb}
\usepackage{bm}
\usepackage{mathtools}
\usepackage{color, xcolor}
\usepackage[linesnumbered,ruled,vlined]{algorithm2e}

\usepackage{overpic}
\usepackage{tikz}

\newcommand\submittedtext{%
	\footnotesize This work has been submitted to the IEEE for possible publication. Copyright may be transferred without notice, after which this version may no longer be accessible.}

\newcommand\submittednotice{%
	\begin{tikzpicture}[remember picture,overlay]
		\node[anchor=south,yshift=10pt] at (current page.south) {\fbox{\parbox{\dimexpr0.65\textwidth-\fboxsep-\fboxrule\relax}{\submittedtext}}};
	\end{tikzpicture}%
}

\begin{document}

\title{AsynEVO: Asynchronous Event-Driven Visual Odometry for Pure Event Streams}

\author{
	
Zhixiang Wang, 
Xudong Li, 
Yizhai Zhang,~\IEEEmembership{Member,~IEEE,} and Panfeng Huang,~\IEEEmembership{Senior Member,~IEEE}
\thanks{This work was supported by the National Natural Science Foundation of China under Grant 62022067.  (\emph{Corresponding author: Yizhai Zhang.})}

\thanks{Zhixiang Wang is with the Research Center for Intelligent Robotics, Shaanxi Province Innovation Team of  Intelligent Robotic Technology, School of Automation, Northwestern Polytechnical University, Xi'an 710072, China (e-mail: wangzhixiang@mail.nwpu.edu.cn).}

\thanks{Xudong Li, Yizhai Zhang, and Panfeng Huang are with the Research Center for Intelligent Robotics, Shaanxi Province Innovation Team of  Intelligent Robotic Technology, School of Astronautics, Northwestern Polytechnical University,  Xi'an 710072, China (e-mail: lxdli@mail.nwpu.edu.cn; zhangyizhai@nwpu.edu.cn; pfhuang@nwpu.edu.cn).}
}

\maketitle

\submittednotice

\begin{abstract}
Event cameras are bio-inspired vision sensors that asynchronously measure per-pixel brightness changes.
The high-temporal resolution and asynchronicity of event cameras offer great potential for estimating robot motion states.
Recent works have adopted the continuous-time estimation methods to exploit the inherent nature of event cameras. 
However, existing methods either have poor runtime performance or neglect the high-temporal resolution of event cameras. 
To alleviate it, an Asynchronous Event-driven Visual Odometry (AsynEVO) based on sparse Gaussian Process (GP) regression is proposed to efficiently infer the motion trajectory from pure event streams. Concretely, an asynchronous frontend pipeline is designed to adapt event-driven feature tracking and manage feature trajectories;  a parallel dynamic sliding-window  backend is presented within the framework of sparse GP regression on $SE(3)$. Notably, a dynamic marginalization strategy is employed to ensure the consistency and sparsity of this GP regression.   Experiments conducted on public datasets and real-world scenarios demonstrate that AsynEVO achieves competitive precision and superior robustness compared to the state-of-the-art.
The experiment in the repeated-texture scenario indicates that the high-temporal resolution of AsynEVO plays a vital role in the estimation of high-speed movement.
Furthermore, we show that the computational efficiency of AsynEVO significantly outperforms the incremental method. 
\end{abstract}


\section{Introduction}


\IEEEPARstart{T}{he} ego-motion estimation plays a critical role in the automated robots \cite{zhang2022monocular, wang2024localization}. 
Typically, the frame-based camera is adopted to infer the ego-motion from intensity images, which is called Visual Odometry (VO) \cite{yu2023tightly, dai2022self} or Simultaneous Localization and Mapping (SLAM) \cite{li2021deepslam, han2024basl}. Over the past decades, an implicit paradigm has formed where a visual frontend tracks the sparse visual features to establish feature associations, and a parallel backend triangulates landmarks from the related features and solves the Maximum A Posteriori (MAP) problem to infer the ego-motion \cite{yu2023tightly, han2024basl}. However, the traditional frame-based feature tracking and discrete-time MAP estimation methods have a limited temporal resolution for the fixed sampling frequency. Moreover, the frame-based camera captures the instantaneous brightness of the current field view at a fixed frequency, and thus the inter-frame motion information is almost abandoned. Although a camera with high frame rates can attenuate the drawback, the resulting computational consumption due to booming images would be quickly intractable. Therefore, the traditional estimation paradigm for frame-based cameras naturally has a limited temporal resolution.

In recent years, event cameras have emerged as a revolutionary technology in the robotics field. Unlike frame-based cameras, event cameras have an asynchronous trigger mechanism that can independently sense the illumination intensity variation at the one-pixel level. This special design brings unique benefits to the event camera, such as high-temporal resolution, High Dynamic Range (HDR), low power consumption, resistance to motion blur\cite{gallego2020event}, etc.  
Nonetheless, it is challenging to realize high-temporal resolution state estimation using the traditional frame-based feature tracking and discrete-time MAP estimation methods, as previous aforementioned reasons. 
Therefore, new frontend tracking and state estimation methods that exert the high-temporal resolution of event cameras are imperative for event-based VO \cite{huang2023event}. 

Asynchronous event-driven feature tracking methods \cite{alzugaray2020haste, gehrig2018asynchronous, alzugaray2018asyn} and Continuous-Time (CT) estimation formulas \cite{mueggler2015continuous, wang2023event, mueggler2018continuous} become a focus of event-based state estimation research recently. 
The event-driven tracker independently tracks each feature in an event-by-event manner, where the detecting and tracking is triggered by asynchronous event streams rather than frame samples. 
Meanwhile, the CT estimation formulas can naturally model the asynchronous measurements. Consequently, the temporal resolution of event cameras is retained as much as possible.
Among existing CT works, two prominent methods have gathered considerable attention, i.e., the B-spline-based methods \cite{mueggler2018continuous} and the Gaussian Process (GP)-based methods \cite{liu2022asynchronous}. 
The former leverages the basic functions to interpolate motion trajectories for the corresponding asynchronous measurements and updates it by adjusting the control points. 
In contrast, the GP-based methods adopt a Gaussian process assumption on partial system states and propagate the probabilistic states with the system state equation. Therefore, they are more flexible and have a clearer physical meaning \cite{tang2019white}. However, the computational complexity of existing GP-based methods for event cameras is substantial, which hinders their practical deployment on robotics \cite{liu2022asynchronous, wang2023event}. 
In this paper, we presents a whole estimation pipeline known as Asynchronous Event-driven Visual Odometry (AsynEVO), consisting of the asynchronous event-driven frontend and dynamic sliding-window CT backend, to infer the motion trajectory from event cameras. The proposed asynchronous event-driven frontend leverages a registration-map mechanism to manage asynchronous feature trajectories efficiently, which retains the high-temporal resolution of event cameras.
A sliding-window optimizer for GP regression on $SE(3)$ is introduced to bound the computational complexity of CT estimator. A dynamic marginalization strategy is developed to maintain the sparsity and consistency of the underlying probability inference problem. 
Our experiments on diverse datasets and real-world scenarios confirm that the proposed approach provides enhanced time efficiency and robustness while maintaining competitive accuracy. In addition, the high-temporal resolution of the proposed method has the advantage of estimating high-speed movement. 
The main contributions of this article are summarized as follows:

\begin{enumerate}{}{}
	\item{An asynchronous event-driven frontend, where the feature trajectory is managed with a registration-map mechanism and the high-temporal resolution of event camera is retained.}
	
	\item{A sliding-window optimizer for GP regression on $SE(3)$, which can naturally model the asynchronous measurements and have a bounded computational complexity. The sparsity and consistency of the underlying factor graph is maintained by the proposed dynamic marginalization strategy.}
	
	\item{The implementation and comprehensive experimental assessments of AsynEVO that leverages the proposed asynchronous data-driven frontend and sliding-window backend to infer motion trajectory from pure event streams. Comparison experiments show that AsynEVO outperforms the state-of-the-art in aspects of time-efficiency, high-speed estimation, and HDR environments.  }
\end{enumerate}

The rest of this article is organized as follows. Sec.~\ref{sec related work} reviews the relevant literature.  In Sec.~\ref{sec methodology}, we first overview the whole system framework, and then introduce the design and model of components, respectively. 
Sec.~\ref{sec experiment} demonstrates the superiority of our proposal on public and own-collected data.  Finally, the conclusions are drawn in Sec.~\ref{sec conclusions}.

\section{Related Work}
\label{sec related work}

Researchers have proposed different schemes to realize ego-motion estimation for event cameras, mainly including frame-like, learning-based, and continuous-time methods. Frame-like methods accumulated event streams to create some composite feature structures, such as Surface of Active Events (SAE) and event frame\cite{rebecq2016evo, wang2023event}. Sparse features were then detected and tracked in a frame-by-frame manner using enhanced conventional feature-based methods \cite{Guan2023pl, Vidal2018Ultimate}. Subsequently, the discrete-time backend optimization or filtering techniques were applied to estimate the motion states. Some studies presented approaches to enhance the frontend performance by directly combining the inertial measurements \cite{rebecq2017real} or other sensors \cite{chen2023esvio}. As the depth of environments can not be measured by monocular camera, event-based stereo visual odometry was also studied to restore the metric scale \cite{zhou2021event, zuo2022devo}. However, frame-like methods primarily treat the event-based estimation problem as a synchronous discrete-time paradigm, which limited their ability to harness the asynchronicity and high-temporal resolution of event cameras. Learning-based methods, including Convolutional Neural Network (CNN) and Spiking Neural Network (SNN), have been designed to predict optical flow from event streams \cite{gehrig2020event, paredes2019unsupervised}. 
A motion estimation module is attached behind the optical flow prediction module to infer the motion states. These CNN-based methods usually process event streams in a synchronous batched manner. Consequently, the asynchronicity of event cameras largely neglected. In contrast, the SNN has an effective mechanism for encoding temporal associations from asynchronous event streams. Some unsupervised learning methods have been suggested to predict both optical flow and ego-motion \cite{zhu2019unsupervised, paredes2019unsupervised}. Typically, these SNN models require  specialized neuromorphic chips for deployment and have inherent difficulty in model training. Moreover, the generalization performance of learning-based methods needs to be improved in future research \cite{gehrig2020event}. 
CT estimation methods \cite{dong2018sparse, wu2022picking, hug2022continuous} were designed to infer the motion trajectory from asynchronous measurements and unsynchronized sensors, such as those taken from a moving car equipped with a scanning radar \cite{wu2022picking} or multi-sensor extrinsic calibration \cite{le2023gaussian}. Recently, CT estimation methods have also been applied to event-based trajectory estimation tasks, as the event camera is inherently an asynchronous sensor. For example, Mueggler \emph{et al}. \cite{mueggler2015continuous, mueggler2018continuous} employed the B-spline to represent the CT trajectory. The estimated motion trajectory is deformed by optimizing the sparse control points to minimize reprojection errors.  

An important component of CT method for event-based motion estimation is based on the GP regression \cite{wang2023event, liu2022asynchronous}. 
Liu \emph{et al}. \cite{liu2022asynchronous} proposed an asynchronous optimization pipeline based on GP regression where the incremental iterative optimizing was triggered by every tracked feature. 
Since each feature is tracked for hundreds of thousands times by the event-driven feature tracker, the computational consumption of optimization pipeline would be quickly intractable. Conversely, we suggest to adopt a sliding-window optimizer enhanced with a dynamic marginalization strategy to improve the computational efficiency of event-based CT motion estimation.
Wang \emph{et al}. \cite{wang2023event} proposed a GP-based system for stereo event cameras. They simultaneously clustered the raw event streams into event frames and SAEs. Then, these features, detected and matched on event frames, will be assigned independent timestamps using the SAEs. Although this is a sound way to realize asynchrony, the inter-frame information is still largely abandoned.
Instead, we design a totally asynchronous event-driven visual frontend that takes advantage of high-temporal resolution of event camera and reserves raw motion information better.

\section{Methodology}
\label{sec methodology}

\subsection{System Overview} \label{sec system overview}

The overall system pipeline primarily consists of two parallel components, i.e., 1) \emph{asynchronous event-driven visual frontend}, and 2) \emph{dynamic sliding-window backend}. In the frontend, the corner features are detected on the SAE. Apart from conventional frame-based methods, the detecting is triggered by the incoming events and is thus asynchronous. When a new feature is detected, we create a data structure, known as feature trajectory, to record the pixel movement of this feature. We attach a separate event-driven tracker to each feature trajectory. The subsequent events, located within the neighbor region, will be tracked in an event-by-event manner. Those feature trajectories that meet preset conditions are fed to the backend. In the backend, several motion states with the same time interval are initialized to form a sliding-window at system startup. Then, we attempt to triangulate asynchronous feature trajectories within the sliding-window. The corresponding state of each measurement on the feature trajectory is queried with the GP-based interpolation. A landmark will be assigned to the feature trajectory if it is successfully triangulated. We further refine all landmarks and states with a full optimization. After initialization, the incoming feature trajectories are triangulated and add to sliding-window sequentially. If the timestamp of the incoming measurement is larger than the counterpart of the latest state, we execute the dynamic marginalization and the sliding-window optimization. Within the sliding-window, the optimization is defined as a finite factor graph where the adjacent motion states are constrained by the GP prior factor and the event measurement is represented as the GP projection factor. 

\subsection{Asynchronous Visual Frontend}\label{sec asynchronous visual frontend}

To realize high-temporal resolution, we design an asynchronous event-driven frontend where sparse features are detected  and tracked in an event-by-event manner. Although separate event-driven feature detecting and tracking methods have been proposed in previous literature, a whole event-driven frontend needs further design on strategies of screening and maintenance. Concretely, we leverage a registration table to record the unique identifier of each active event feature. The pixel location of each event feature can be infer from its index in the registration table. Then, the unique identifier is mapped to the corresponding feature trajectory and its independent tracker. When a event measurement is obtained, we first search in the registration table to determine if its neighbor region already exists activate event features. If there is at least one active event feature, we utilize the identifier-tracker map to find the corresponding tracker and feed the event measurement to it. The movement of active feature will be reported by the tracker and collected into the feature trajectory. 
Otherwise, we update the SAE and determine if an eHarris corner feature can be detected at the pixel location of this event measurement. 

Since the  registration table has the same width and height as the resolution of event cameras, it is high-efficient to access and modify elements in the registration table. The search of existing features only pays attention to a small patch within the registration table, and thus has a low time complexity. The small patch also reject the feature clustering and realize sparse features. Similarly, the detection only occurs at the location of the current event measurement and is thus efficient. As the tracking of active features is triggered by subsequent events, a special case is that a active feature have not be triggered for a long time. For example, a corner feature is detected in noisy events by mistake, and thus it should be discarded promptly. To realize it, we assume the active feature should be triggered frequently within a maximum time threshold. We periodically delete the active features that have not been updated for more than the maximum time threshold. In addition, we also set a minimum time threshold to avoid the feature trajectory collecting a mass of redundant measurements. With patch search and time threshold mechanisms, the result frontend can maintain sparse and asynchronous feature tracking.

\begin{figure}[!t]
	\centering
	\includegraphics[width=3.3in]{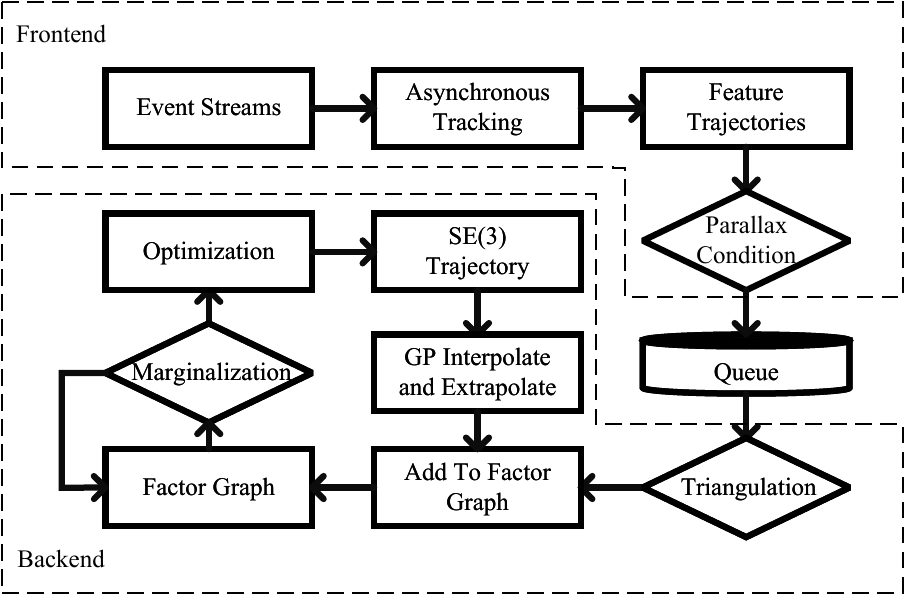}
	\caption[System pipeline.]{System pipeline. The first thread inputs event measurements, tracks with asynchronous tracking method, and outputs feature trajectories to a queue. The second thread receives feature trajectories, tries triangulating them, and adds them into the backend graph.}
	\label{system_pipeline}
\end{figure}

\subsection{Backend Problem Description}\label{sec problem description}
The feature trajectory is defined as
\begin{equation}
	\label{ft}
	\boldsymbol{F}_{i} \triangleq \langle  \boldsymbol{m}_1^i,\boldsymbol{m}_2^i,...,\boldsymbol{m}_n^i \rangle ,
\end{equation}
which represents an ordered spatial-temporal association set. The $\boldsymbol{m}_{\tau}^i$ indicates the $\tau$-th tracking of the landmark $\boldsymbol{l}_i$, and it  contains 
\begin{equation}
	\label{fts_pixel_position}
	\boldsymbol{m}^{i}_{\tau} \triangleq \{ t_{i, \tau}, \boldsymbol{q}_{i,\tau} \},
\end{equation}
where $t_{i, \tau}$ is the measurement timestamp and $\boldsymbol{q}_{i,\tau} = [x_{i, \tau}, y_{i, \tau}]^{\top}$ is the pixel position.

Suppose feature trajectories  are collected from $M$ landmarks as the camera moves according to a certain CT trajectory $\boldsymbol{\chi}$ in scenario. Let $\boldsymbol{F}$ be the set of feature trajectories, where $\boldsymbol{F} = \{\boldsymbol{F}_{i} | i \in \{0, 1, \dots, M-1\}\}$. In GP regression, the CT trajectory $\boldsymbol{\chi}$ is parameterized as sample points $\boldsymbol{x}_k\ (k \in \{ 0, 1, 2, \dots, N \})$. 
Given the measurement time $t_{i,\tau}$  as defined in \eqref{fts_pixel_position}, the observation state $\boldsymbol{x}(t_{i, \tau})$ can be queried on $\boldsymbol{\chi}$ using its nearest states $\boldsymbol{x}_{k}$ and $\boldsymbol{x}_{k+1}$, where  $t_{k} \leq t_{i,\tau} < t_{k+1}$. 
The accumulative Mahalanobis norm of GP prior residuals and GP projection measurement residuals are minimized to obtain a maximum posterior estimation, which is represented as
\begin{equation}
	\begin{split}
		\label{OP}
		\mathop{\min}_{\boldsymbol{x}, \boldsymbol{l}}\sum_{k=0}^{N-1} ( &\Vert e(\boldsymbol{x}_k,\boldsymbol{x}_{k+1}) \Vert_{\boldsymbol{Q}_{k+1}}^2 \\
		&+ \sum_{\boldsymbol{F}_{i} \in \boldsymbol{F}, \boldsymbol{m}^i_{\tau}\in \boldsymbol{F}_{i}}\Vert e(\boldsymbol{x}_k,\boldsymbol{x}_{k+1},\boldsymbol{l}_i) \Vert_{\boldsymbol{R}}^2),
	\end{split}
\end{equation}
where $e(\boldsymbol{x}_k,\boldsymbol{x}_{k+1})$ indicates the GP prior residual, $ e(\boldsymbol{x}_k,\boldsymbol{x}_{k+1},\boldsymbol{l}_i)$ represents the GP projection residuals,  and $\boldsymbol{Q}_{k+1}$ and $\boldsymbol{R} $ present the covariance matrices.

\begin{figure}[!t]
	\centering
	\includegraphics[width=3.4in]{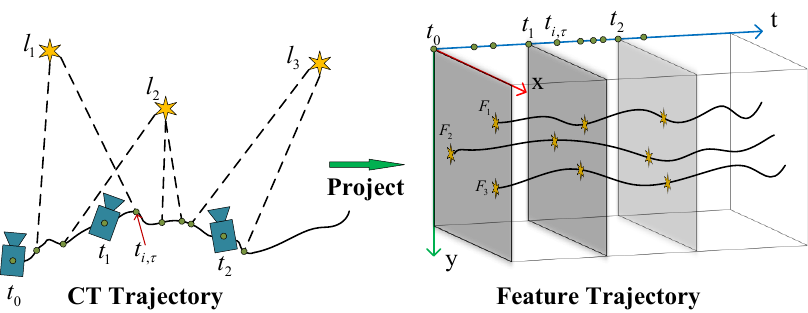}
	\caption[Illustration of feature trajectories.]{Illustration of feature trajectories. When the event camera observes some landmarks and moves in the scenario, the event stream triggered by the same landmark will be managed as a feature trajectory.}
	\label{feature trajectory illustration}
\end{figure}

\subsection{Sparse GP priors on $SE(3)$} \label{sec Sparse GP priors}
Define the motion state as $\boldsymbol{x}(t) \triangleq \{\boldsymbol{T}(t),\boldsymbol{\varpi}(t)\}$,  where $\boldsymbol{T}(t)$ represents the matrix that transforms a vector from the body frame to the world frame, and $\boldsymbol{\varpi}(t)$ denotes the generalized velocity in the body frame.
Assume White-Noise-On-Acceleration (WNOA)  prior \cite{tang2019white}, and the GP-based kinematics model is presented as
\begin{align}
	\label{GPprior without linear}
	\dot{\boldsymbol{T}}(t) &= \boldsymbol{T}(t) (\boldsymbol{\varpi}(t))^\land \nonumber \\
	\dot{\boldsymbol{\varpi}}(t) &= \boldsymbol{w}(t) \nonumber \\
	\quad \boldsymbol{w}(t) &\sim \boldsymbol{GP}(\mathbf{0},\boldsymbol{Q}_c \delta(t-t')),
\end{align}
where $(\bullet)^{\land}$ maps a vector in $\mathbb{R}^{6}$ to $\mathfrak{se}(3)$ \cite{barfoot2017state}, $\dot{\boldsymbol{\varpi}}(t)$ is the generalized acceleration vector, $\boldsymbol{Q}_{c}$ is a usual power spectral density matrix, and $\delta(t-t')$ is Dirac's delta
function.
Since $\boldsymbol{x}(t) \in SE(3) \times \mathbb{R}^{6}$ forms a manifold, it is difficult to directly deploy this nonlinear model.  To linearize \eqref{GPprior without linear},  the local variable $\boldsymbol{\xi}$ in the tangent space of $\boldsymbol{T}_k$ is given by
\begin{equation}
	\label{Tlinear}
	\boldsymbol{T}(t) = \boldsymbol{T}_k \exp(\boldsymbol{\xi}_k(t)^\land),
\end{equation}
where $\exp(\bullet)$ is the exponential map from $\mathfrak{se}(3)$ to $SE(3)$. 
Actually, the  derivative of $\boldsymbol{\xi}_k(t)$ can serve as the body-frame velocity in the local frame. In fact, we have 
\begin{equation}
	\label{rellocalvelocity}
	\boldsymbol{\varpi}(t) = \boldsymbol{J}_r (\boldsymbol{\xi}_k(t))\dot{\boldsymbol{\xi}}_k(t)
\end{equation}
to describe relationship between the actual body-frame velocity and the linearized velocity, where $\boldsymbol{J}_r(\boldsymbol{\xi}_k(t))$ represents the right Jacobian of $\boldsymbol{\xi}_k(t)$. Similarly, the local linear state can be defined by
\begin{equation}
	\label{localstate}
	\boldsymbol{\gamma}_k(t) \triangleq \left[
	\begin{array}{cc}
		\boldsymbol{\xi}_k(t)\\
		\dot{\boldsymbol{\xi}}_k(t)
	\end{array}
	\right].
\end{equation}
Then, the local linearized GP kinematics model can be written as
\begin{equation}
	\label{GPprior with linear}
	\dot{\boldsymbol{\gamma}}_k(t) = \left[
	\begin{array}{cc}
		\dot{\boldsymbol{\xi}}_k(t)\\
		\ddot{\boldsymbol{\xi}}_k(t)
	\end{array}
	\right], \quad \ddot{\boldsymbol{\xi}}_k(t)=\boldsymbol{w}(t) \sim \boldsymbol{GP}(0,\boldsymbol{Q}_c\delta(t-t')).
\end{equation}
Known the local linear state $\boldsymbol{\gamma}_k(t)$, we can use \eqref{Tlinear} and \eqref{rellocalvelocity} to recover the global motion state as
\begin{equation}
	\label{Recover}
	\boldsymbol{x}(t) = \{\boldsymbol{T}_k \exp(\boldsymbol{\xi}_k(t)^\land),\boldsymbol{J}_r (\boldsymbol{\xi}_k(t))\dot{\boldsymbol{\xi}}_k(t)\}.
\end{equation}
Under the WNOA assumption, the GP prior residual between timestamp $t_k$ and $t_{k+1}$ can be defined as
\begin{equation}
	\label{PriorRes}
	e(\boldsymbol{x}_k,\boldsymbol{x}_{k+1}) = \left[\begin{array}{cc}
		(t_{k+1}-t_k)\boldsymbol{\varpi}_k- \log(\boldsymbol{T}_k^{-1}\boldsymbol{T}_{k+1})^\vee\\
		\boldsymbol{\varpi}_k-\boldsymbol{J}_r (\log(\boldsymbol{T}_k^{-1}\boldsymbol{T}_{k+1})^\vee)^{-1}\boldsymbol{\varpi}_{k+1}
	\end{array}
	\right],
\end{equation}
where $\log(\bullet)$ and $(\bullet)^{\vee}$ are the inverse maps of $\exp(\bullet)$ and $(\bullet)^{\land}$, respectively. 

\subsection{GP Projection Measurement}\label{sec visual measurement model}
From the foregoing, we need to query the camera state $\boldsymbol{x}(t_{i,\tau})$ for each asynchronous visual measurement $\boldsymbol{m}_{\tau}^i$ on feature trajectories.
We firstly apply the GP assumption in linear space, and the local state $\boldsymbol{\gamma}(t_{i,\tau})$ can be interpolated as  
\begin{align}
	\label{Query}
	\boldsymbol{\gamma}_k(t_{i,\tau})&=\boldsymbol{\Lambda}(t_{i,\tau})\boldsymbol{\gamma}_k(t_k) + \boldsymbol{\Psi}(t_{i,\tau})\boldsymbol{\gamma}_k(t_{k+1}) \nonumber \\
	\boldsymbol{\Lambda}(t_{i,\tau})&=\boldsymbol{\Phi}(t_{i,\tau},t_k)-\boldsymbol{Q}_{i,\tau} \boldsymbol{\Phi}(t_{k+1},t_{i,\tau})^{\top}\boldsymbol{Q}_{k+1}^{-1}\boldsymbol{\Phi}(t_{k+1},t_k) \nonumber \\
	\boldsymbol{\Psi}(t_{i,\tau})&=\boldsymbol{Q}_{i,\tau} \boldsymbol{\Phi}(t_{k+1},t_{i,\tau})^{\top}\boldsymbol{Q}_{k+1}^{-1},
\end{align}
where $\boldsymbol{\Phi}(t_{\tau},t_k)$ is the transition matrix \cite{dong2018sparse} and $\boldsymbol{Q}_{i,\tau}$ is obtained through substitution of the measurement timestamp $t_{i,\tau}$ into the GP prior covariance matrix \eqref{Qp}. 
Then, we can use \eqref{Recover} to recover $\boldsymbol{x}(t_{i,\tau})$. Therefore, the projected pixel position of landmark $\boldsymbol{l}_{i} = [x_{i}, y_{i}, z_{i}]^{\top}$ at timestamp $t_{i,\tau}$ can be written as
\begin{equation}
	\label{Measur}
	\hat{\boldsymbol{q}}_{i,\tau} = \frac{1}{d_{i}} \boldsymbol{P} \boldsymbol{K} \boldsymbol{T}(t_{i,\tau})^{\top} \boldsymbol{l}_{i} + \boldsymbol{\epsilon},
\end{equation} 
where $\boldsymbol{K}$ indicates the intrinsic matrix, $\boldsymbol{P}$ transforms a vector from 3D to 2D, $d_{i}$ is the depth of this landmark in current camera frame, and $\boldsymbol{\epsilon} \sim \mathcal{N}(\mathbf{0},\boldsymbol{R})$ represents the Gaussian pixel noise. Then, the measurement residual in \eqref{OP} is defined by
\begin{equation}
	\label{MeasurRes}
	e(\boldsymbol{x}_k,\boldsymbol{x}_{k+1}, \boldsymbol{l}_{i}) = \hat{\boldsymbol{q}}_{i,\tau} - \boldsymbol{q}_{i,\tau}.
\end{equation}

\subsection{Dynamic Sliding-Window} \label{sec dynamic sliding-Window}

Existing CT estimation methods generally employ the incremental smoothing (or full smoothing \cite{mueggler2018continuous}) backend to tackle optimization problems. However, these methods suffer from sensitivity to outliers and are computational inefficient. Inspired by discrete-time state estimation methods, we introduced an approach similar to the fixed-lag smoother \cite{lv2023continuous}. More specifically, we establish a sliding-window framework and subsequently employ dynamic marginalization to eliminate redundant states. Diverging from the fixed-lag smoother, we marginalize states by considering the interconnections among feature trajectories and motion states, instead of maintaining a fixed number of keyframes. 

\begin{figure*}[!t]
	\centering
	\includegraphics[width=7in]{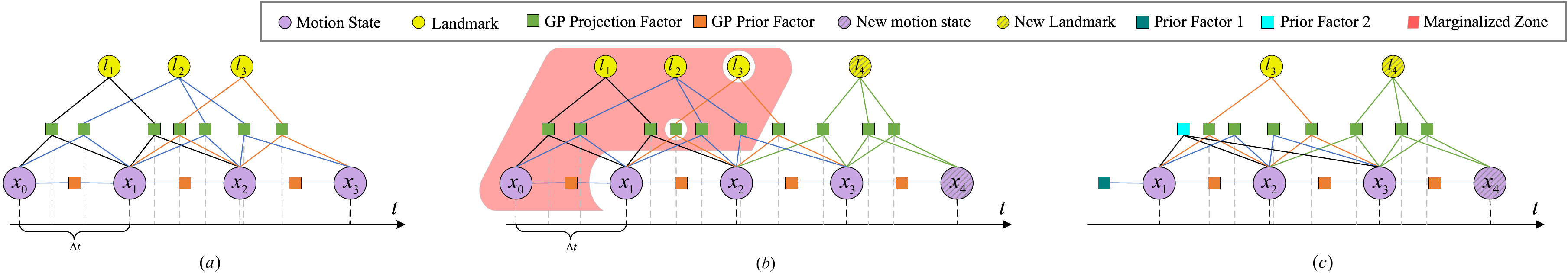}
	\caption[Factor graph in dynamic sliding-window.]{Factor graph in dynamic sliding-window. (a) The original factor graph. (b) The factor graph that add new landmarks and states. The marginalized zone is marked with red shade. Notice that the node surrounded by a white ring inside the marginalized zone is not marginalized. (c) Marginalized factor graph. }
	\label{Factor graph in dynamic sliding-window}
\end{figure*}

\begin{algorithm}
	\caption{Dynamic Marginalization}
	\label{alg DMS}
	\KwData{Motion trajectory $ \boldsymbol{\chi}$; \newline Feature trajectory set $\boldsymbol{F}$; \newline Minimum window size $N_{min}$; \newline
		Map function $\mathcal{F}(\boldsymbol{x}_{k}) \rightarrow \boldsymbol{F}_{k,r}$. 
	}
	\KwResult{Motion states $\boldsymbol{\chi}_{m}$ to be marginalized; \newline  Feature trajectories $\boldsymbol{F}_{m}$ to be marginalized.}
	$\boldsymbol{\chi}_{m} \leftarrow \emptyset$\;
	$\boldsymbol{F}_{m} \leftarrow \emptyset$\;
	\For{$\boldsymbol{F}_{i}$ \textbf{in} $\boldsymbol{F}$}{
		$\boldsymbol{m}_{1}^{i} = \boldsymbol{F}_{i}.front()$\;
		$\boldsymbol{m}_{n}^{i} = \boldsymbol{F}_{i}.back()$\;
		$t_{front} = t_{i,1}, \quad where \ t_{i,1} \in \boldsymbol{m}_{1}^{i}$\;
		$t_{back} = t_{i,n}, \quad where \ t_{i,n} \in \boldsymbol{m}_{n}^{i}$\;
		$t_{0} = \boldsymbol{x}_{0}.t, \quad where \ \boldsymbol{x}_{0} \in \boldsymbol{\chi}$\;
		$t_{1} = \boldsymbol{x}_{1}.t, \quad where \ \boldsymbol{x}_{1} \in \boldsymbol{\chi}$\;
		$t_{N} = \boldsymbol{x}_{N}.t, \quad where \ \boldsymbol{x}_{N} \in \boldsymbol{\chi}$\;
		$t_{\epsilon} = 0.2 t_{0} + 0.8t_{N}$\;
		\If{$t_{back} < t_{\epsilon}$ and $t_{0} \leq t_{front} <t_{1}$}{
			$\boldsymbol{F}_{m} \leftarrow \boldsymbol{F}_{m} \cup \{\boldsymbol{F}_{i}\}$\;
		}
	}
	\For{$\boldsymbol{x}_{k}$ \textbf{in} $\boldsymbol{\chi}$}{
		\tcp{$N$ is current window size}
		\If{$ N < N_{min} $}{\textbf{break}\;}  
		$\boldsymbol{F}_{k, r} \leftarrow \mathcal{F}(\boldsymbol{x}_{k})$\;
		\If{$\boldsymbol{F}_{k, r} == \emptyset$}{
			$\boldsymbol{\chi}_{m} \leftarrow \boldsymbol{\chi}_{m} \cup \{\boldsymbol{x}_{k}\} $\;
		}
		
		\If{$\boldsymbol{F}_{k,r} \neq \emptyset$}{
			\textbf{break}\;
		}
	}
	$\boldsymbol{\chi}  \leftarrow \boldsymbol{\chi} - \boldsymbol{\chi}_{m}$\;
	$\boldsymbol{F} \leftarrow \boldsymbol{F} - \boldsymbol{F}_{m}$.
\end{algorithm}

The pseudocode of the proposed marginalization algorithm is presented in Alg.~\ref{alg DMS}. The function $\mathcal{F}(\boldsymbol{x}_{k}) \rightarrow \boldsymbol{F}_{k,r}$ maps a motion state $\boldsymbol{x}_{k}$ and its next neighbor $\boldsymbol{x}_{k+1}$ to their corresponding feature trajectory set $\boldsymbol{F}_{k,r}$, where  $\boldsymbol{F}_{k,r} \subseteq \boldsymbol{F}$. 
These feature trajectories, created between the initial two motion states within the sliding window and terminated before the latest $1/5$ of motion states, are considered to be marginalized (Line 3-13 in Alg.~\ref{alg DMS}).
Simultaneously, a motion state is marginalized if it is at the current beginning of the sliding-window and all associated feature trajectories have already been marked for marginalization (Line 14-21 in Alg.~\ref{alg DMS}). Furthermore, a minimum sliding-window is guaranteed to prevent excessive marginalization (Line 15-16 in Alg.~\ref{alg DMS}).

The special sparse structure of bundle adjustments plays a pivotal role in maintaining  computational efficiency \cite{dellaert2017factor}. With our marginalization algorithm, we neither discard any measurements directly nor introduce new constraints between motion states and landmarks. Therefore, the structure of the resulting Hessian matrix retains its arrow-shaped form, implying that the corresponding optimization problem still have a distinct sparse structure and could be solved efficiently. Since the resulting sliding-window can vary with the current data-association condition, we refer to this approach as a \emph{dynamic sliding-window} or \emph{dynamic marginalization}. Based on the factor graph theory, the marginalization is achieved on the linearized Markov blanket of the state variables \cite{usenko2019visual}. Thus, the Markov blanket is first extracted from the original factor graph according to the state variables that should be marginalized, i.e., the outputs of the Alg.~\ref{alg DMS}. Then, the linearization of Markov blanket is achieved by the Schur complement \cite{barfoot2017state}. Finally, the induced marginal distribution on the Markov blanket replaces the corresponding part in the original factor graph as presented in Fig.~\ref{Factor graph in dynamic sliding-window}.

\begin{figure}[!t]
	\centering
	\includegraphics[width=3.3in]{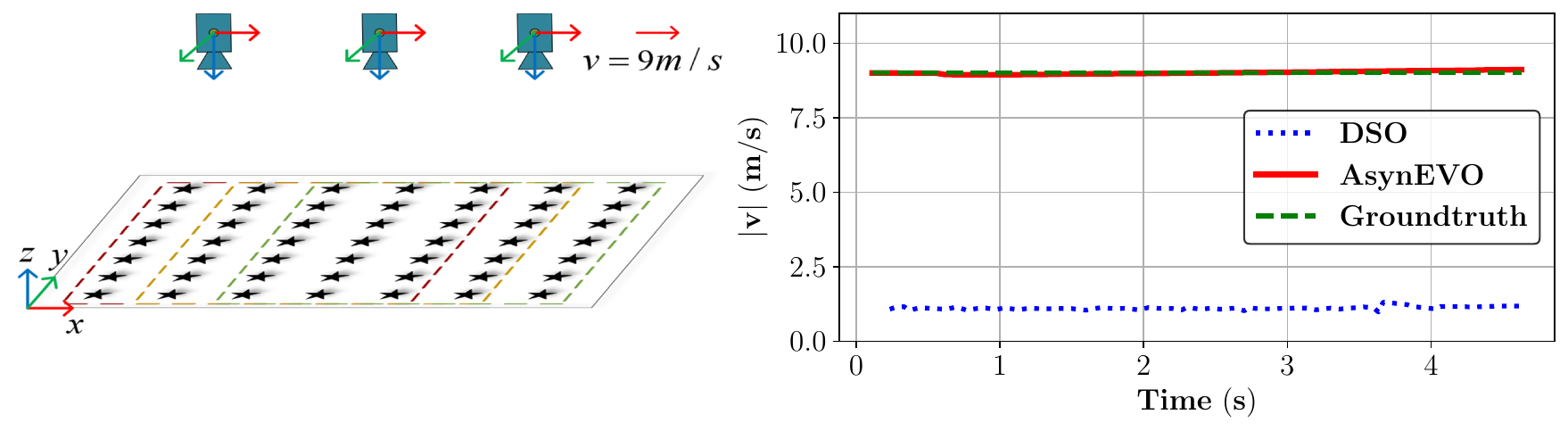}
	\caption[Repeated-texture scenario and estimation result.]{Repeated-texture scenario and estimation result. The event camera translates in $9\ m/s$. The AsynEVO can estimate the true motion correctly while the frame-based visual odometry gets a wrong result for its limited temporal resolution. Note that the velocity of DSO is calculated by differencing neighbor poses. }
	\label{asyn_vs_syn}
\end{figure}

\section{Experiments}
\label{sec experiment}

\subsection{Implementation Details}

The proposed pipeline is implemented in C++ and  GTSAM\footnote{https://github.com/borglab/gtsam}. The GP prior factors and GP projection factors is built according to \cite{dong2018sparse}. 
In the asynchronous frontend, the feature detecting is based on FA-Harris \cite{li2019fa} and the feature tracking is realized with Haste \cite{alzugaray2020haste}. 
Our evaluations are conducted on a standard computer with Intel Xeon Gold 6226R @ 3.90 GHz, Ubuntu 20.04 and ROS Noetic.

\subsection{Asynchronicity vs Synchronicity}
An intuitive example is reported to demonstrate the advantage of asynchronicity over synchronicity. The event camera quickly moves in a repeated-texture scenario. The gray images and event streams are simultaneously collected by the ESIM \cite{rebecq2018esim}. Then, we estimate the camera motion trajectory by a conventional frame-based methods called Direct Sparse Odometry (DSO) \cite{engel2017direct} and our AsynEVO, respectively.  Since the gray images are captured in fixed frequency ($30\ Hz$), the repeated-texture will cause illusion if the camera translates in a certain velocity. As shown in Fig.~\ref{asyn_vs_syn}, the estimated states of DSO has a lower speed rather than the ground truth.  However, our proposal can still infer the ground truth velocity, as shown in Fig.~\ref{asyn_vs_syn}. The reason of this phenomenon is that our proposal can leverage the high-temporal resolution of event camera better. It demonstrates the necessity of designing the asynchronous event-driven estimator as realized in this paper.

\subsection{Precision and Robustness}

\begin{figure}[!t]
	\centering
	\includegraphics[width=3.5in]{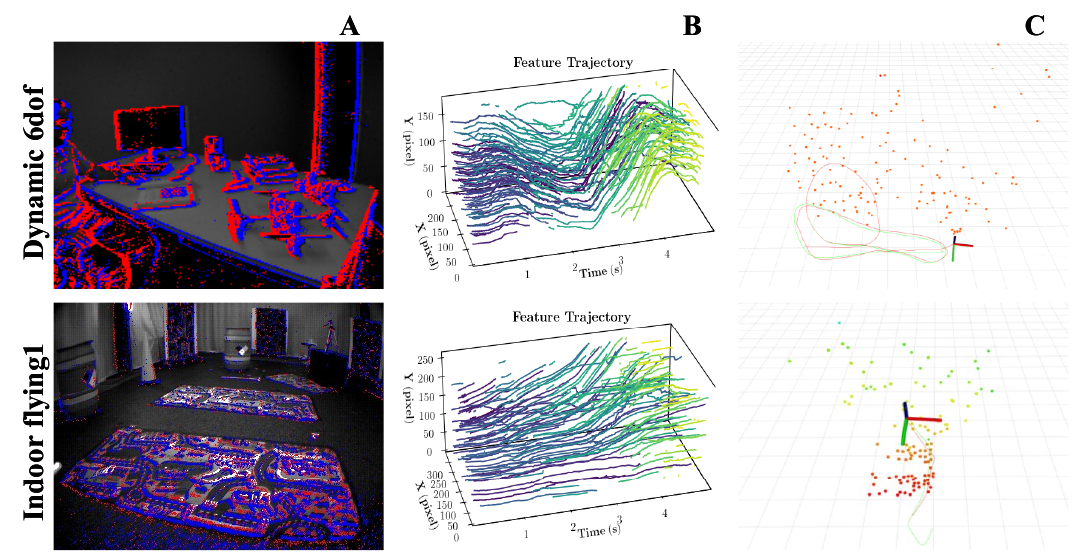}
	\caption[Experimental scenarios and intermediate results.]{Experimental scenarios and intermediate results. Scenarios and event measurements are visualized in (\textbf{A}). Dynamic 6dof and Indoor flying1 come from public datasets which are captured by DAVIS event cameras. (\textbf{B}) Sparse feature points are detected and tracked by the proposed frontend, which forms a mass of feature trajectories in 3D time-pixel-plane coordinates. The estimated motion trajectories and corresponding ground-truth are illustrated in (\textbf{C}) The pointcloud as an intermediate product is also shown in (\textbf{C}).}
	\label{Experiment_Result}
\end{figure}

\begin{figure}[!t]
	\centering
	\includegraphics[width=3.5in]{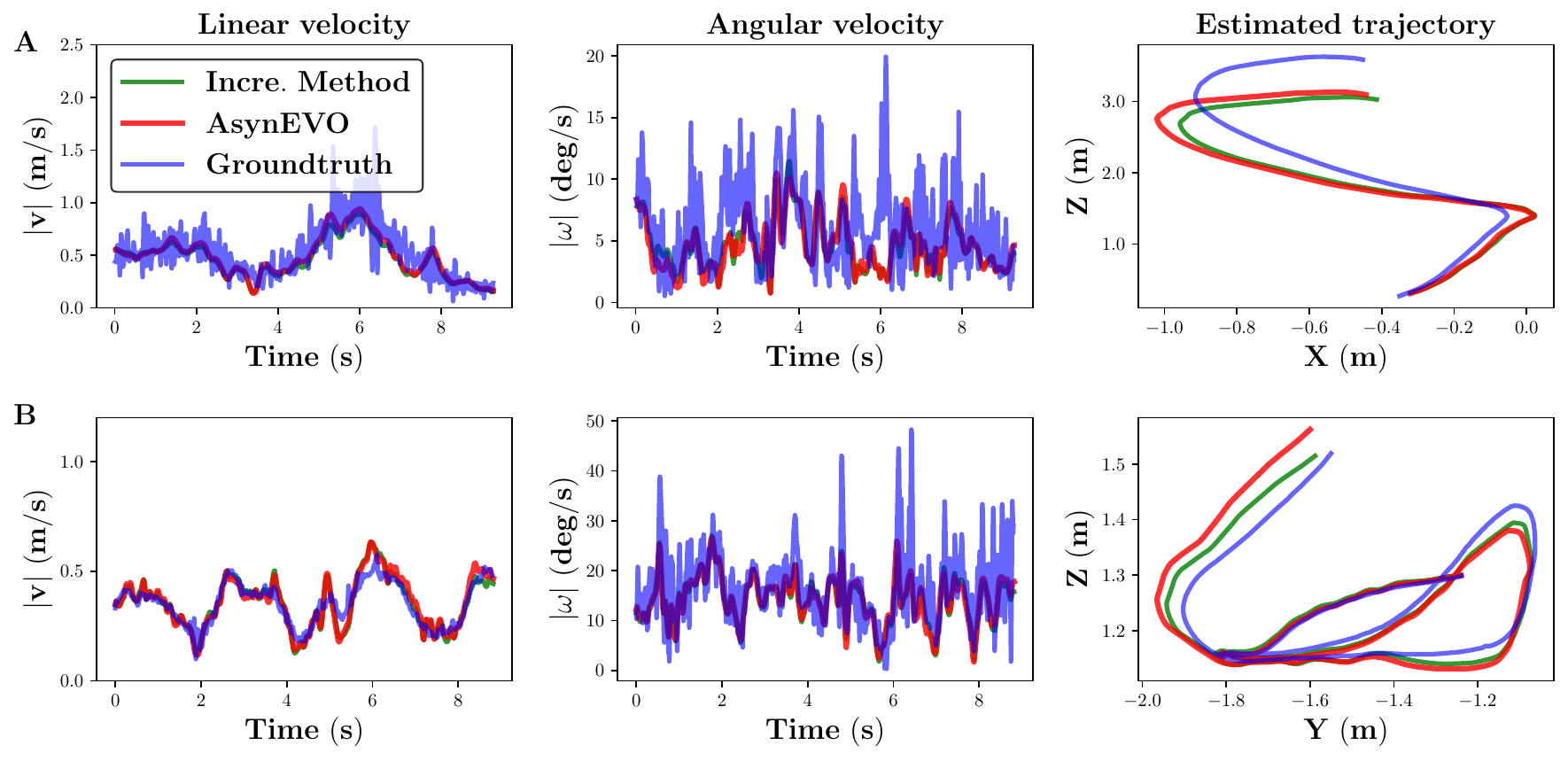}
	\caption[Estimation results on public datasets.]{Estimation results on public datasets. (\textbf{A}) Indoor flying1. (\textbf{B}) Dynamic 6dof. Both linear and angular velocities are basically consistent with the ground truth. The noisy measurements in (\textbf{A}) and (\textbf{B}) are directly captured by a monocular DAVIS camera. As a result, the estimated trajectories are disturbed by the accumulation drift and the scale drift. Compared with time-consuming incremental methods, our lightweight AsynEVO can achieve similar estimation accuracy.  }
	\label{estimate error}
\end{figure}

To evaluate the precision and robustness of the proposed pipeline, we test it on both the public datasets and the own-collected datasets. The public datasets include the DAVIS 240C dataset \cite{mueggler2017event} and the MVSEC dataset \cite{zhu2018multivehicle} that are generated by DAVIS cameras. The DAVIS 240C ($240 \times 180$ pixels) dataset consists of  intensity images and ground truth from a motion capture system. The MVSEC dataset is collected using a stereo camera (DAVIS 346B, $346 \times 260$ pixels). However, we only use the left event streams and ground truth for assessment.

To compare with the state-of-the-art method, the asynchronous optimization pipeline proposed in \cite{liu2022asynchronous} is replicated. For fairness, we adopt the same asynchronous frontend and realize the same system initialization as our proposal. The significant difference is the incremental backend that is triggered asynchronously. Otherwise, these two methods are also compared with the DSO \cite{engel2017direct} to show the difference. All estimated motion trajectories are aligned in $SIM(3)$ with the ground truth to address scale drifts.
The rest of parameters are consistent with our method during the comparison.

Experimental scenarios are shown in Fig.~\ref{Experiment_Result}\textbf{A}. Meanwhile, the feature trajectories are visualized in  Fig.~\ref{Experiment_Result}\textbf{B}, where each color represents an individual feature trajectory. The asynchronicity can be illustrated by the arbitrary start time and end time of each feature trajectory. The estimated 3D trajectories (red line) and  their ground truth (green line) are represented in Fig.~\ref{Experiment_Result}\textbf{C}. The proposed  monocular pipeline is confused by inherent scale drift. Nonetheless, the shape of the estimated trajectory almost coincides with its ground truth. In addition, the landmarks, estimated as intermediate products, are also displayed in Fig.~\ref{Experiment_Result}\textbf{C}, which nearly coincides with their corresponding feature tracking results. These experiments demonstrate that the AsynEVO is effective to track asynchronous feature trajectories and estimate motion trajectories from event streams.

The curves of velocities and positions are visualized in Fig.~\ref{estimate error}. Intuitively, the precision of the incremental method is competitive with AsynEVO. Note that we adopt the same visual frontend as AsynEVO to the incremental method for fair comparisons. Therefore, the precision of incremental method also benefits from our event-driven asynchronous frontend.
The Root Mean Square Relative Trajectory Errors (RMS RTE) are shown in Table.~\ref{RMSE_table}. The DSO has realized more accurate estimations on most datasets. However, the DSO receives low  accuracy and fails in the HDR scenarios (hdr boxes and hdr poster), which demonstrates that the event-based VO has advantages in some difficult scenario for traditional VO methods. Our proposal performs better than the incremental method on 7 among 12 sequences. 
As the AsynEVO marginalizes partial state variables during the estimation, the estimation errors in the previous state variables could be accumulated gradually and degenerate the precision of subsequent state variables. In contrast, the incremental approach retains all state variables and measurement information.
In the evaluation, we found that the incremental method is sensitive to outliers and linear system errors. To complete the evaluation, the outliers are removed from the factor graph of the incremental method. 
In contrast, the dynamic sliding-window pipeline can successfully run on real-world datasets without the outlier exclusion, indicating that our method is more robust than the incremental method. The reason is that the AsynEVO utilizes the Levenberg-Marquardt method to solve the least squares optimization problem. In the Levenberg-Marquardt method, a scaled identity matrix is added to the original Hessian matrix to guarantee the non-singularity of the linear system.

\begin{table*}[!t]
	\begin{center}
		\caption[The RMS RTE on public event datasets.]{The RMS RTE on various event datasets.}
		\label{RMSE_table}
		\begin{tabular}{ c | c | c | c | c | c | c}
			\hline\hline
			Method  &      dynamic 6dof & shapes 6dof & shapes translation & poster 6dof & poster translation & boxes 6dof \\
			\hline
			
			DSO &    0.005 & 0.009 & 0.007 & 0.009 & 0.080 & 0.006  \\
			
			Incre. method &     0.019 & 0.119 & 0.079 & 0.060 & 0.039 & 0.059  \\ 
			
			\textbf{AsynEVO} &     0.027 & \textbf{0.113} & \textbf{0.038} & 0.075 & \textbf{0.027} & 0.067 \\
			\hline
			\hline
			
			Method &  boxes translation & hdr boxes & hdr poster & indoor flying1 & indoor flying2 &  indoor flying3 \\
			\hline
			
			DSO & 0.006 & 0.299 & - &0.075 & 0.109 &  0.565 \\
			Incre. method & 0.069 & 0.050 & 0.026  & 0.097 & 0.218 &  0.207 \\
			\textbf{AsynEVO} & \textbf{0.022} & 0.070 & 0.035  & \textbf{0.090} & \textbf{0.213} &  \textbf{0.177}  \\
			\hline
			
		\end{tabular}
	\end{center}
\end{table*}

\begin{figure}[!t]
	\centering
	\includegraphics[width=3.4in]{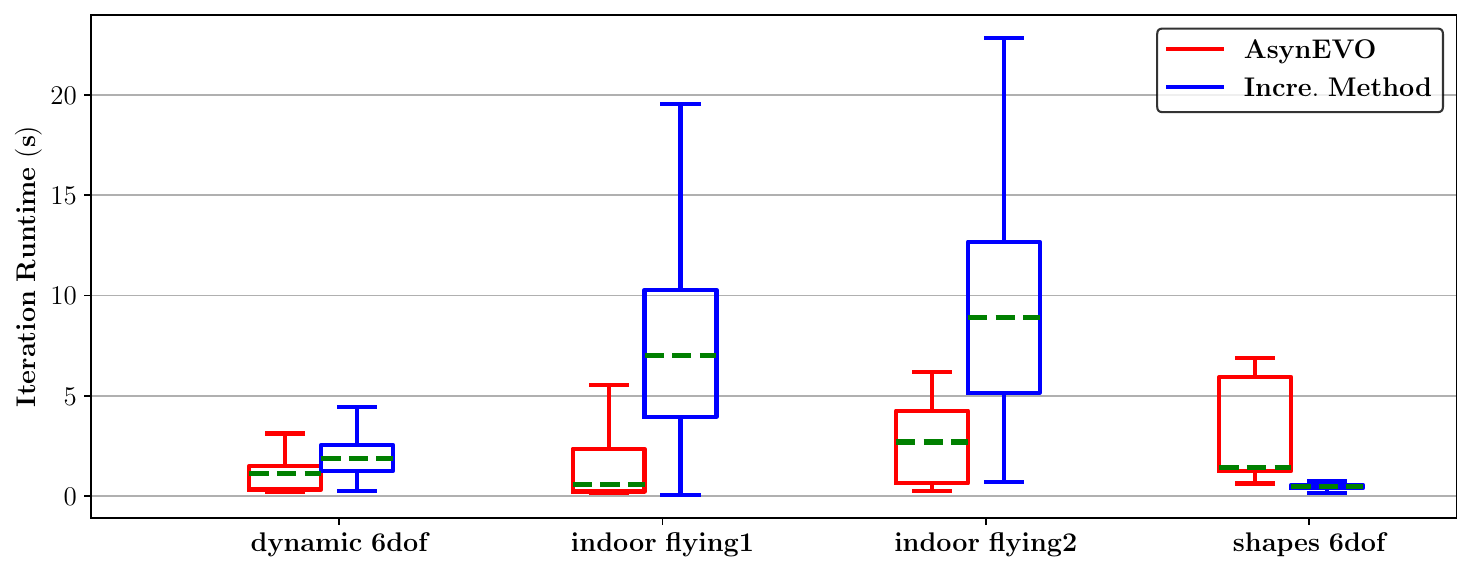}
	\caption[Runtime of AsynEVO compared with the incremental method.]{Runtime of AsynEVO compared with the incremental method. On most datasets, the proposal has a prominent reduction in runtime.}
	\label{computational_time_boxplt}
\end{figure}

\subsection{Computational Consumption}

\begin{figure}[!t]
	\centering
	\includegraphics[width=3.4in]{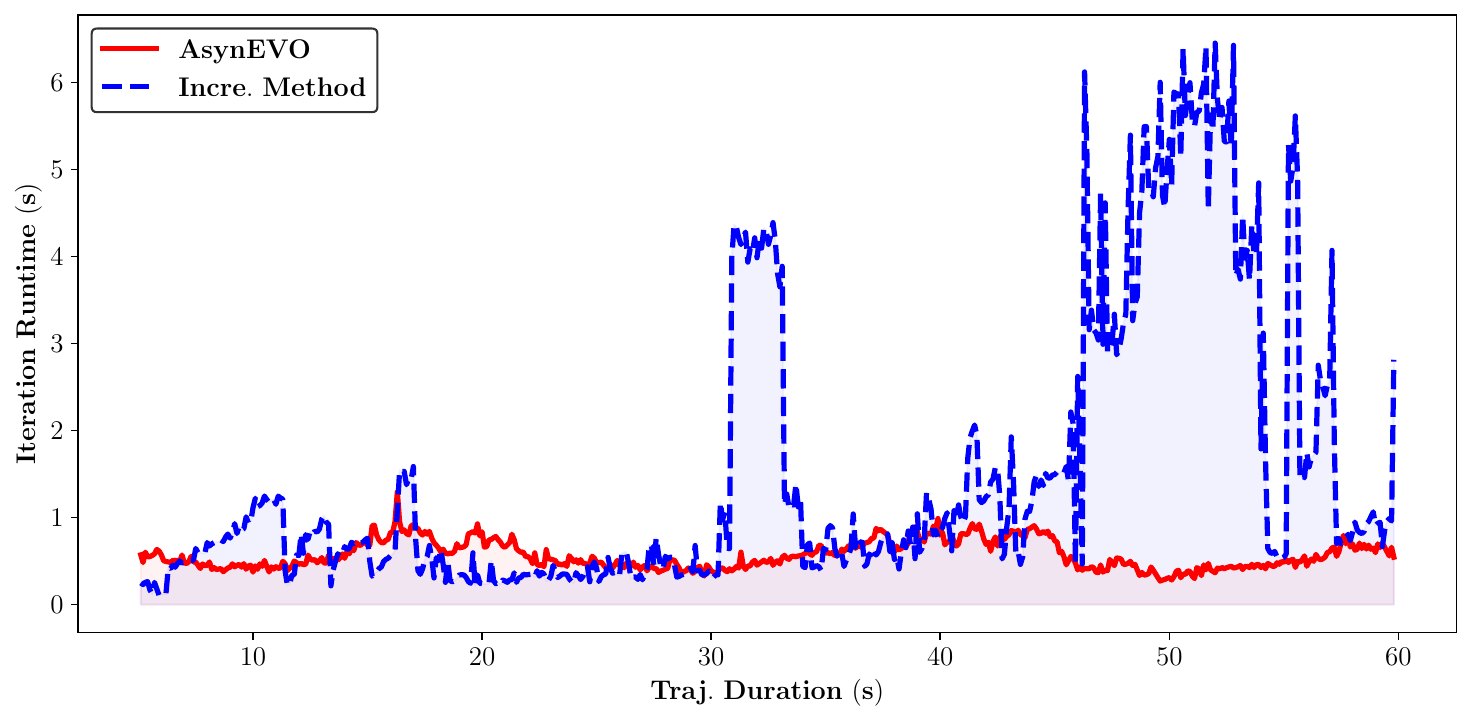}
	\caption[Runtime on the whole trajectory.]{Runtime on the whole trajectory. Compared with the proposal, the incremental method has a higher and significantly unstable runtime.}
	\label{computational_time}
\end{figure}

The iteration runtime on part datasets is visualized in Fig.~\ref{computational_time_boxplt}. Our AsynEVO has a significant improvement in estimation efficiency, and yet the runtime of the incremental method is very unstable (as shown in Fig.~\ref{computational_time}). On \emph{shapes 6dof} sequence, the polygon shapes almost never move out of field view, so the measurements and states are never marginalized, which results in a higher runtime cost. 
On most datasets, the optimizer of AsynEVO maintains a limited scale by the sliding-window and marginalization. As the fixed-time updating is much more important for real robot's state estimation, the AsynEVO has more potential to be used in future practical scenarios.

\begin{figure}[!t]
	\centering
	\includegraphics[width=3.3in]{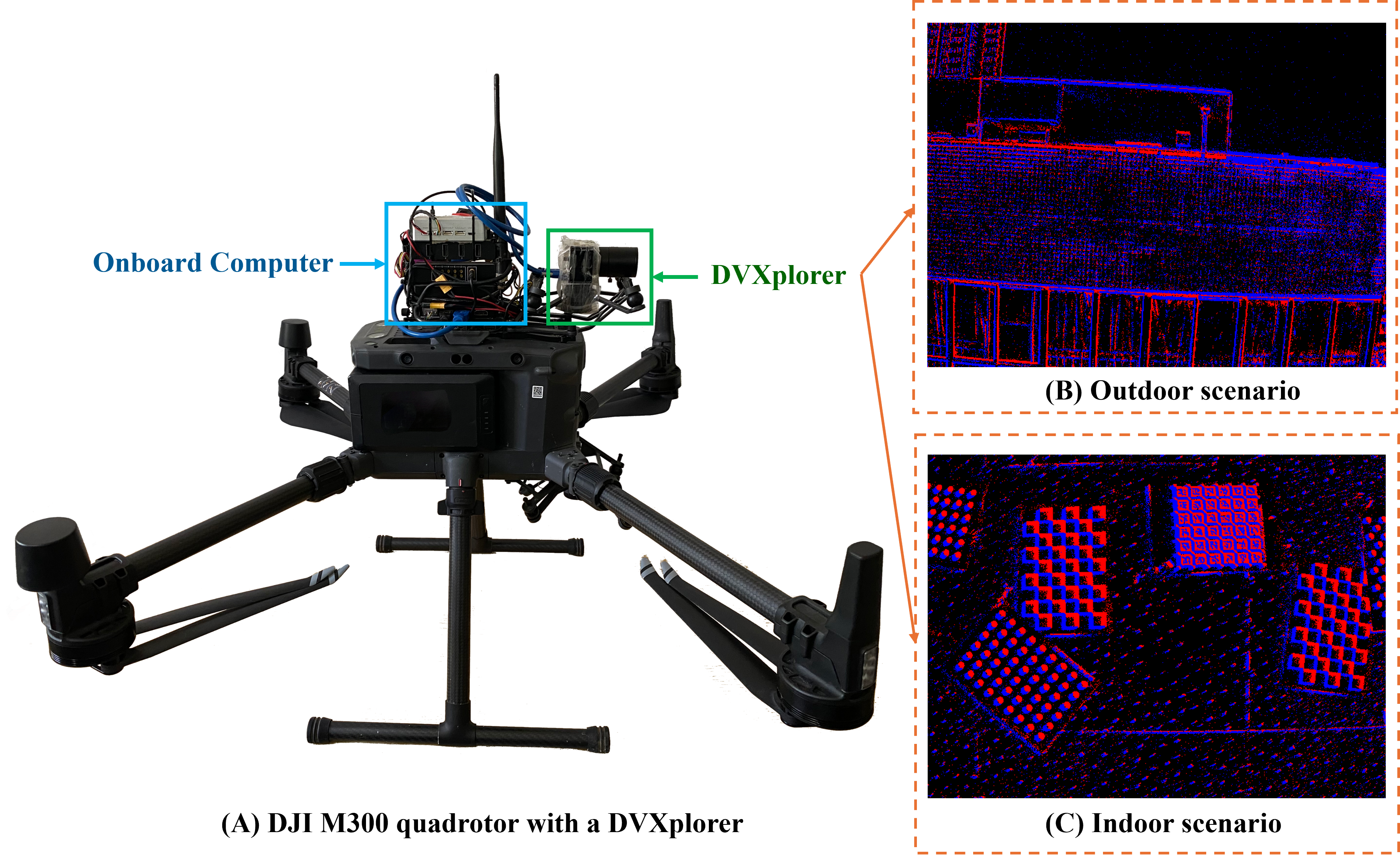}
	\caption[Quadrotor and scenarios used in real-world experiments.]{Quadrotor and scenarios used in real-world experiments. The DVXplorer can not capture intensity images, and AsynEVO can infer the motion trajectory from pure event streams.}
	\label{uav_equip}
\end{figure}

\begin{figure}[!t]
	\centering
	\includegraphics[width=3.5in]{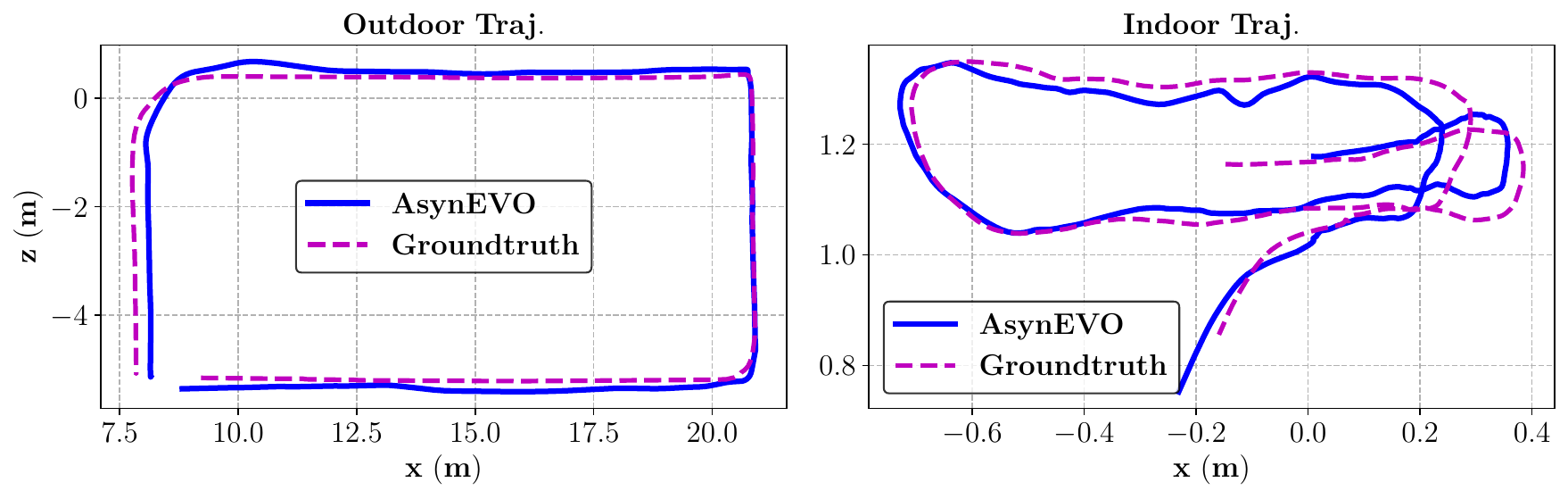}
	\caption[Estimation results of AsynEVO on own-collected datasets.]{Estimation results of AsynEVO on own-collected datasets.}
	\label{fig:own_collected}
\end{figure}

\subsection{Real-World Experiments}
The real-world experiments are deployed to collect the event datasets with a DVXplorer ($640 \times 480$ pixels) event camera. Unlike the DAVIS camera, the DVXplorer only capture pure event streams without gray images. We equip the DVXplorer on a DJI M300-RTK quadrotor (as shown in Fig.~\ref{uav_equip}) and record the ground truth with the onboard RTK on outdoor flights \cite{wang2024localization}.  In indoor experiments, the estimated trajectories are compared with the ground truth from the OptiTrack.

Benefiting from the high quality event streams of DVXplorer, our AsynEVO can estimation the true motion trajectories in both indoor and outdoor environments. We find that the estimated result of the outdoor flight is generally more accurate than the indoor handheld movement (as shown in Fig.~\ref{fig:own_collected}). The reason might be that the outdoor flight trajectory is more coincident with the WNOA prior than the handheld movement. It means that AsynEVO can receive better estimations when the motion trajectory has smaller accelerations and jerks.

\section{Conclusions}
\label{sec conclusions}

In this paper, an asynchronous visual odometry known as AsynVO was proposed to estimate $SE(3)$ trajectory from asynchronous event streams.  The proposed method was tested to be credible on both public datasets and real-world experiments. Although our current version can't accomplish real-time runtime, the proposed method exhibits potential capacity for future improvements. Since our backend optimization problem becomes partial dense among related poses after the landmark marginalization operation, we can mitigate that by sparsifying the problem through the removal of redundant connections of state variables. Due to the Markov property of the motion state variables, the aforementioned sparsity would not compromise the original optimization problem.
Furthermore, our approach can also be readily extended to integrate stereo, inertial, and White-Noise-On-Jerk (WNOJ) configurations in the future, which would be beneficial to inhibiting the scale drift and supporting aggressive trajectories. The additional observations would permit smaller window size and realize real-time estimation.


\bibliographystyle{IEEEtran}
\bibliography{IEEEabrv,asynodometry}

\end{document}